\documentclass[conference]{IEEEtran}
\usepackage{amsmath,amssymb,amsfonts}
\usepackage{algorithmic}
\usepackage{graphicx}
\usepackage{textcomp}
\usepackage{tablefootnote}
\usepackage{xcolor}
\usepackage{multirow} 
\usepackage{adjustbox}
\usepackage{algorithm}

\def\BibTeX{{\rm B\kern-.05em{\sc i\kern-.025em b}\kern-.08em
    T\kern-.1667em\lower.7ex\hbox{E}\kern-.125emX}}
    
\begin{document}

\title{DWT-CompCNN: Deep Image Classification Network for High Throughput JPEG 2000 Compressed Documents}
\author{
\IEEEauthorblockN{Tejasvee Bisen$^{1}$, Mohammed Javed$^{1}$, Shashank Kirtania$^{2}$, P. Nagabhushan $^{1}$}

\IEEEauthorblockA{
\textsuperscript{1}Department of IT, Indian Institute of Information Technology Allahabad, India \\
\textsuperscript{2} Department of CE, Thapar Institute of Engineering and Technology, India \\
%Emails:\{mit2021079@iiita.ac.in, javed@iiita.ac.in, doermann@buffalo.edu\}\\
\text{Email:\{rsi2018001@iiita.ac.in, javed@iiita.ac.in, shashankkirtania123@gmail.com, pnagabhushan@iiita.ac.in\}}}
}


\maketitle
\begin{abstract}
For any digital application with document images such as retrieval, the classification of document images becomes an essential stage. Conventionally for the purpose, the full versions of the documents, that is the uncompressed document images make the input dataset, which poses a threat due to the big volume required to accommodate the full versions of the documents. Therefore, it would be novel, if the same classification task could be accomplished directly (with some partial decompression) with the compressed representation of documents in order to make the whole process computationally more efficient. In this research work, a novel deep learning model - DWT-CompCNN is proposed for classification of documents that are compressed using High Throughput JPEG 2000 (HTJ2K) algorithm. The proposed DWT-CompCNN comprises of five convolutional layers with filter sizes of 16, 32, 64, 128, and 256 consecutively for each increasing layer to improve learning from the wavelet coefficients extracted from the compressed images. Experiments are performed on two benchmark datasets- Tobacco-3482 and RVL-CDIP, which demonstrate that the proposed model is time and space efficient, and also achieves a better classification accuracy in compressed domain.
\end{abstract}

\begin{IEEEkeywords}
DWT-CompCNN, Document Classification, Compressed Document version, JPEG 2000, Discrete Wavelet Transform, Deep learning in Compressed Domain
\end{IEEEkeywords}


\maketitle

\section{Introduction}\label{sec1}
Classification of document images based on their content and structure plays a significant role in the field of Document Image Analysis (DIA), which drives various useful applications like Optical Character Recognition (OCR), Natural Language Processing (NLP), and archival and retrieval of large documents. In the DIA literature, there are lot of efforts seen to classify document images by designing handcrafted features\cite{kumar2014structural}, \cite{csurka2016right} and more recently by deep learning models \cite{harley2015evaluation}, \cite{sarkhel2019deterministic}. Moreover, most of these reported works are specifically meant to work with uncompressed or raw version of document images; but the general trend nowadays is to compress the acquired images before archival or transmission, to make them space and time efficient \cite{barni2018document}, \cite{zhang2018document}. As a result, huge volume of compressed images are being achieved and transmitted on daily basis over the internet world and digital databases, and therefore developing novel algorithms to operate/analyse these compressed images without involving decompression is the need of the hour. In this direction, there are some recent attempts to develop algorithms that can directly operate with compressed images avoiding the stages of decompression and recompression, and thus offering computational and memory related advantages as reported by \cite{byju2020remote},\cite{mukhopadhyay2011image} and \cite{javed2018review}. Therefore, the present research paper is focused on demonstrating the idea of accomplishing document classification directly using the compressed stream of High Throughput JPEG 2000 (HTJ2K) document images.  
The problem of document classification has been addressed in the DIA literature based on contents and structure, using conventional methods \cite{kumar2014structural}, \cite{csurka2016right} and deep learning methods \cite{afzal2017cutting}, \cite{ferrando2020improving}. Content-based classification algorithms are designed with the help of Optical Character Recognition (OCR) that extract and recognize text information. On the other hand, structure based methods directly use the document contents and layout to classify document images, like in \cite{hu2019document}, MSCNN has been used which extracts features from images using a cross-connect structure and multi-channel convolution network. Classification of document images based on fisher vector and run length histogram is addressed in \cite{csurka2017document}. A deep learning networks for feature extraction and extreme learning machine (ELM) to perform the task of classification is discussed in \cite{kolsch2017real}.
Deep Convolutional Neural Network (DCNN) has been used in the field of document image classification by \cite{das2018document} using inter-domain transfer learning for training the model on the whole document, and intra-domain transfer learning for region-based learning of documents, finally combining both techniques for classification.

In the literature, textual and visual features of document images have also been used combiningly to perform classification, like in \cite{mandivarapu2021efficient}, Graph Convolution Neural Network (GCNN) extracts textual, layout, and visual features and assigns them as nodes of GCNN for further classification. In \cite{bakkali2020visual} and \cite{asim2019two}, textual and visual features of images are extracted using CNN for classification. In \cite{forman2003extensive} textual features are extracted using Tesseract OCR, then features are filtered using Balanced Accuracy Measure (ACC2) algorithm , visual features are extracted using InceptionV3 \cite{szegedy2016rethinking}, both features are embedded using average ensembling algorithm to classify the images in different predefined classes. In \cite{bakkali2020cross}, using the NASNet-Large model for extracting visual features and BERT for textual features, cross-modal approach is devised for classifying document images. Machine learning techniques such as random forest classifier \cite{kumar2014structural}, logistic regression \cite{kang2014convolutional} have been used to extract features of images and classify with help of CNN models. Many works have been done by making variations in existing standard CNN architectures to get better results on different datasets by training the models accordingly \cite{tensmeyer2017analysis}. Motivated from the fact that deep learning models have performed much better classification in the recent years, as underscored above, in this paper, a deep learning based model is conceived for document classification in the compressed domain. 

There are different compression algorithms designed in the literature to deal with document images \cite{salomon2004data}, such as Huffman coding, LZW compression, Run length encoding (RLE) \cite{nagabhushan2014entropy}, Joint Photographic Experts Group (JPEG) that uses Discrete Cosine Transform (DCT) \cite{de1998processing}, Joint Photographic Experts Group 2000 (JPEG2000) that uses Discrete Wavelet Transform  (DWT) \cite{rabbani2002jpeg2000}. Based on different compression algorithms, few researchers have attempted to process images directly in the respective compressed representations such as RLE \cite{nagabhushan2014entropy}, DWT in JPEG2000 and DCT in JPEG for applications like text-line segmentation\cite{nagabhushan2019text}, word recognition\cite{rajesh2021hh}, classification\cite{byju2020remote}, image retrieval \cite{byju2020progressive}\cite{schaefer2017fast}. In \cite{rajesh2019dct} and \cite{arslan2022usage} DCT coefficients of JPEG compressed images have been used to classify the images using deep learning models. Many works have been reported in the wavelet domain, where the wavelet coefficients are extracted using forward transform and then used for different applications like texture classification and document script identification. In \cite{hiremath2008wavelet} and \cite{williams2016advanced} features are extracted from wavelet subbands of images for classification using deep learning models. In \cite{khatami2020convolutional}, \cite{ali2018deep}, and \cite{rossetto2019improving} medical images classification is performed by extracting features of images from wavelet domain. In \cite{li2000context} wavelet coefficients of high frequency subbands and Laplacian distribution have been used to classify document images in four different classes background, text, images, text, and graphs. In \cite{li2020wavelet} wavelets are integrated with CNNs for better performance of models while classifying the images. In\cite{shankar2007neuro} 2D DWT has been used for feature extraction and then fed to the Multilayer perceptron (MLP) to classify the remote sensing images. Most of the above works are reported in wavelet domain, however, in this research paper the idea of wavelet compressed domain is proposed, where the wavelet coefficients are extracted at different resolutions directly from the compressed stream of images and then used for document classification. There are few attempts to work directly in wavelet compressed domain of JPEG2000 where , DWT coefficients extracted from the compressed stream are used for classification of images for network applications \cite{chamain2020improving}, and remote sensing images\cite{byju2020remote}.
However, to the best of our knowledge, the present research work is the first attempt to accomplish document classification directly in the wavelet compressed domain using deep learning model at different resolutions of images.

This paper aims to classify document images directly from the compressed DWT multiresolution features of HTJ2K, using a deep CNN model called as DWT-CompCNN. The proposed approach saves  significant amount of time and space required during decompression and recompression for performing the task of classification as in case of uncompressed images, and also giving better accuracy than uncompressed domain models. The proposed approach works on feature of multiresolution of HTJ2K which makes the proposed model powerful. The experiments are conducted on two standard datasets, Tobacco-3482 and RVL-CDIP. 

Following are the major research contribution in the paper:
\begin{itemize}
    \item DWT-CompCNN network for classification of HTJ2K compressed documents at different resolutions of DWT.
    \item State-of the-art accuracy of 92.04\% and 98.94\% reported for compressed Tobacco-3482 and RVL-CDIP datasets respectively in compressed domain at resolution 3.
    \item The proposed model accomplishes document classification in compressed domain with max speedup of 4.81 at resolution 1 and and min speedup of 1.90 at resolution 3, also with 50\% less memory requirement on Tobacco-3482 dataset.
\end{itemize}
The rest of the paper is organized into five sections: Section II describes the problem background which includes fundamentals of DWT, working of JPEG2000 and High Throughput JPEG 2000 (HTJ2K). Section III explains the proposed approach and deep architecture of DWT-CompCNN for document classification. Section IV reports the experimental results with the proposed DWT-CompCNN model. Section V summarises the present work with some future directions.

\section{Problem Background}
In this section, Discrete Wavelet Transform used in JPEG2000 algorithm, along with the procedure of JPEG2000 and HTJ2K are discussed.
\subsection{Discrete Wavelet Transform (DWT)}
Image compression plays a significant role in image transmission and storage, which aims at minimizing the size of an image without losing substantial information or image quality. Some of the popular image compression techniques are Run length encoding compression (RLE) \cite{abdmouleh2012new}, Discrete Cosine Transform (DCT) \cite{watson1994image}, Discrete Wavelet Transform (DWT) \cite{salomon2004data} and \cite{chowdhury2012image}.
A DWT, decomposes a signal into different subbands, where each subband consists of a time series of coefficients. It describes the signal's time evolution with the respective frequency band. The DWT for the given signal $f$ is described in Equation (\ref{eq1}), where $q$ and $r$ are scaling factor and transformation respectively, $\phi$ is DWT (mother wavelet) for the given signal, which is transformed to be the signal $F$.

\begin{equation}
   F_{q,r} = \int_{-\infty}^{\infty} f(u) \phi_{q,r}(u) du 
   \label{eq1}
\end{equation}

The exploration of the multiresolution analysis led to the development of wavelets theory. It also provides significant information about the relation between time and frequency on several resolutions. The Equations (\ref{eq2}) and (\ref{eq3}) represent the decomposition of an image at two resolution levels by applying a 2D wavelet transform and Equation (\ref{eq4}) helps to get back the original image using Inverse Discrete Wavelet Transform (IDWT). 
For an image $I(m\times n)$ with dimension $X\times Y$, and the two dimensional DWT is defined below in the Equation (\ref{eq2}) for the given image $I$:
\begin{equation}
    A_\chi (L_{o}, x,y) = \frac{1}{\sqrt{X.Y}} \sum_{m=0}^{X-1} \sum_{n=0}^{Y-1} I(m,n) \chi_{L_{o},x,y} (m,n)
    \label{eq2}
\end{equation}
\begin{equation}
    A_\Psi (L, x,y) ^{j} = \frac{1}{\sqrt{X.Y}} \sum_{m=0}^{X-1} \sum_{n=0}^{Y-1} I(m,n) \Psi^{j}_{L,x,y} (m,n)
    \label{eq3}
\end{equation}
The inverse wavelet transform is defined as:\\
\begin{equation}
\begin{split}
     I(m \mathrm{x} n) =  \frac{1}{\sqrt{X.Y}} \sum_{m} \sum_{n} I(m,n)  A_\chi (L_{o}, x,y) \chi_{L_{o},x,y} (m,n) \\ + \frac{1}{\sqrt{X.Y}} \sum_{j=h,v,d} \sum_{L= L_{o}}^{\infty}\sum_{m} \sum_{n}  A_\Psi (L, x,y) ^{j} \Psi^{j}_{L,x,y} (m,n)
\end{split}
\label{eq4}
\end{equation}
where $A_\chi$ and $A_\Psi$ are approximation and detail coefficients respectively, subband dimensions are \emph{(m,n)}, also \emph{j} is subband set for h (Horizontal), v (Vertical), d (Diagonal) and resolution level is $L$, and $L_{o}$ is highest level resolution. In this paper, multiresolution features of DWT is used for classification in compressed domain.
\begin{figure*}[h]
	%\begin{center}
		\centering{\includegraphics[width=12cm,height=10cm]{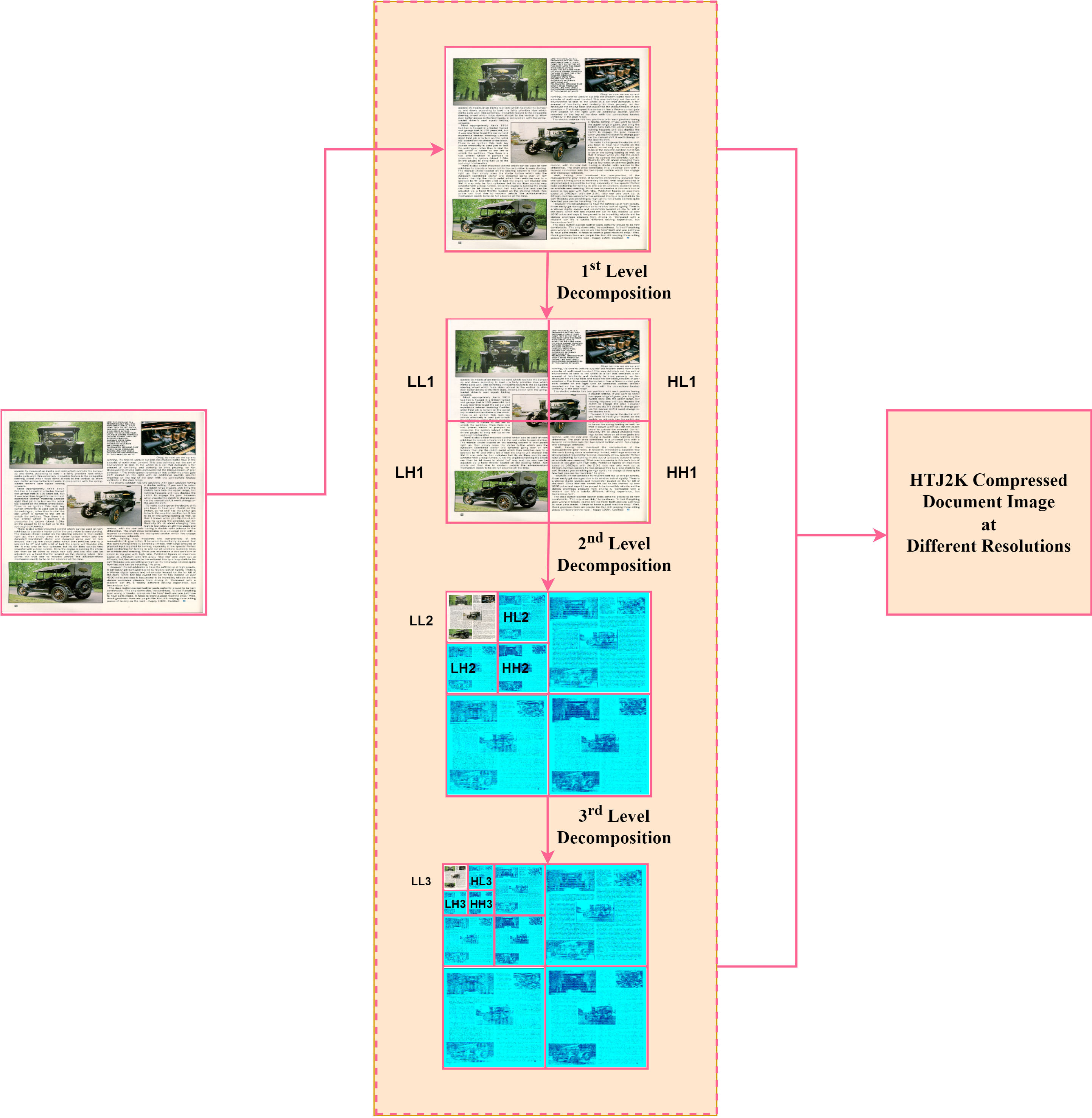}} %\vspace*{8pt}
		\caption{ Three level decomposition illustrated for Discrete Wavelet Transform (DWT) used in HTJ2K, where LL subband consist the approximate image, HL subband stores horizontal features, HL subband stores the vertical features, LH subband stores the diagonal features of the image.}
		\label{3d_dwt}
	%\end{center}
\end{figure*}
\subsection{JPEG 2000}
JPEG 2000\cite{schelkens2009jpeg} is one of the compression algorithms, which uses DWT as a compression technique. JPEG 2000 contains components for compression and decompression of image, which are  DWT, quantization and EBCOT encoding. DWT helps to decompose the image at different resolutions. The Fig.\ref{3d_dwt} shows the three level DWT decomposition of images used by JPEG 2000. In quantization step, the wavelet coefficients are quantized, for lossless compression of images, quantization on images are not performed. After performing DWT on images, it is divided into different subbands which are further divided into blocks known as precincts, which are again subdivided into code blocks. The code blocks are coded using Embedded Block Coding with an Optimized Truncation algorithm (EBCOT). Embedded block coding, has two tiers, Tier-1 is a step in which code blocks are coded using arithmetic coding, and further in Tier-2 bitstreams are obtained after arithmetic coding is organized into packets.

JPEG2000's multiresolution feature is being used by the HTJ2K for making it time and space efficient. The HTJ2K algorithm is used by the proposed approach to partially decompress images at different resolutions. 
Wavelet based features are retrieved from the partially decompressed archives, such as shape based features, and texture from approximation information of LL subband from different resolutions, and are used for classification.\\
\begin{figure*}[h]
	\begin{center}
		\centering{\includegraphics[width=11.5cm,height=6cm]{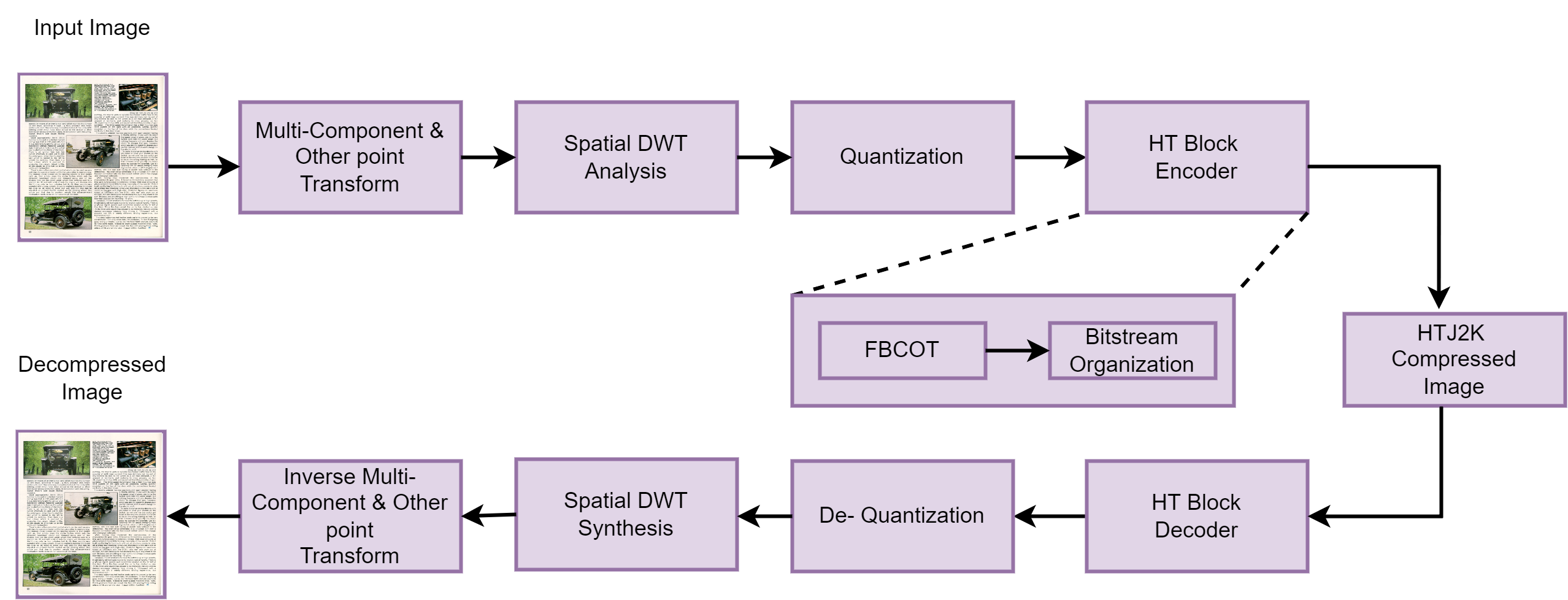}} 
		\caption{ HTJ2K compression and decompression block diagram}
		\label{jpeg 2000}
	\end{center}
\end{figure*}

\subsection{High Throughput JPEG 2000 (HTJ2K)}
High Throughput JPEG 2000 \cite{taubman2019high} is JPEG 2000's Part 15 of standard JPEG 2000, which overcomes the computational complexity of Embedded Block Coding with Optimized Truncation (EBCOT) used as the block coding algorithm in JPEG 2000. The block structure of HTJ2K is illustrated in Fig.\ref{jpeg 2000}, which shows the compression and decompression of HTJ2K. The HTJ2K uses Fast Block Coding with Optimized Truncation (FBCOT) as a block coding unit, which reduces the iterative encoding as used by EBCOT to reduce the computational complexity.

The JPEG 2000's block coding algorithm Embedded Block Coding with Optimized Truncation (EBCOT) uses three coding passes, Cleanup (CUP), Significance Propagation (SP), and magnitude Refinement (MR) respectively. Each CUP pass enhances the quality of all samples in the code-block, but the SP and MR refinement runs simply enhance the quality of specific samples. The approach is computationally demanding since JPEG 2000 Part-1 (J2K1) performs three trips over the code-block for each bit-plane. This approach used by JPEG 2000 is computationally expensive as it performs three passes over the code-block for each bit-plane. The HT block coding uses coding passes as used by JPEG 2000, but is specified in terms of bit-planes. The CUP pass of HT block coding encodes the magnitude and signs for each bit-planes. The rest of the two passes of HT coding, SP, and MR are the same as the original JPEG 2000. The HT block decoder just requires to process one CUP, one SP, and one MR pass to fully retrieve the information, however, a JPEG 2000 block decoder may need to process a large number of passes as compared to the HT block decoder. Hence HT block compression system helps to make the system computationally effective. In this research paper, the efficient HTJ2K is used for classification in the compressed domain.

\begin{figure*}[h]
	%\begin{center}
		\centering{\includegraphics[width=12cm,height=10cm]{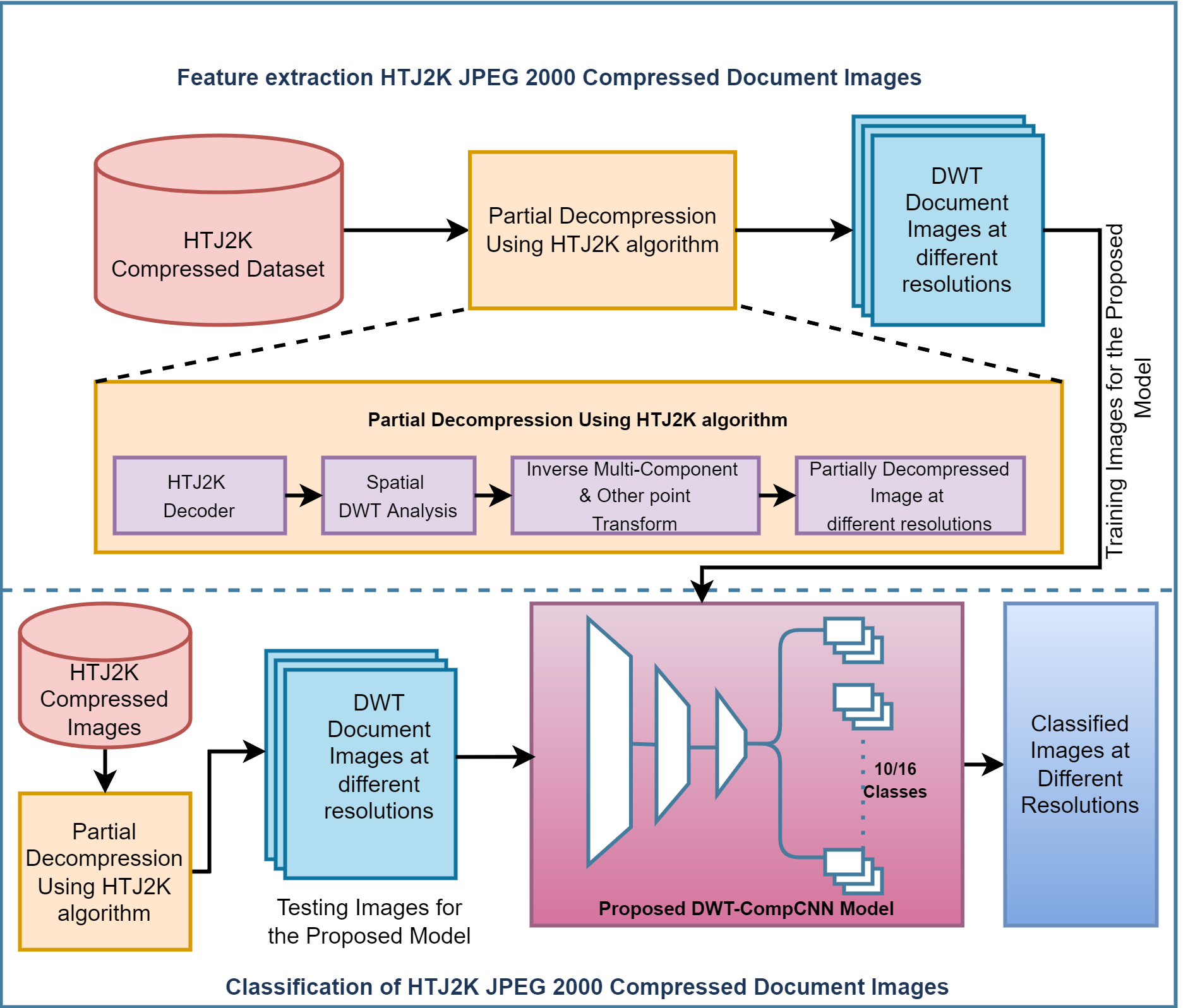}} \vspace*{8pt}
		\caption{ Block diagram of the proposed approach with DWT-CompCNN model for classification of document images in HTJ2K compressed domain}
		\label{proposed model}
	%\end{center}
\end{figure*}

\begin{figure*}[h]
	%\begin{center}
		\centering{\includegraphics[width=12cm,height=10cm]{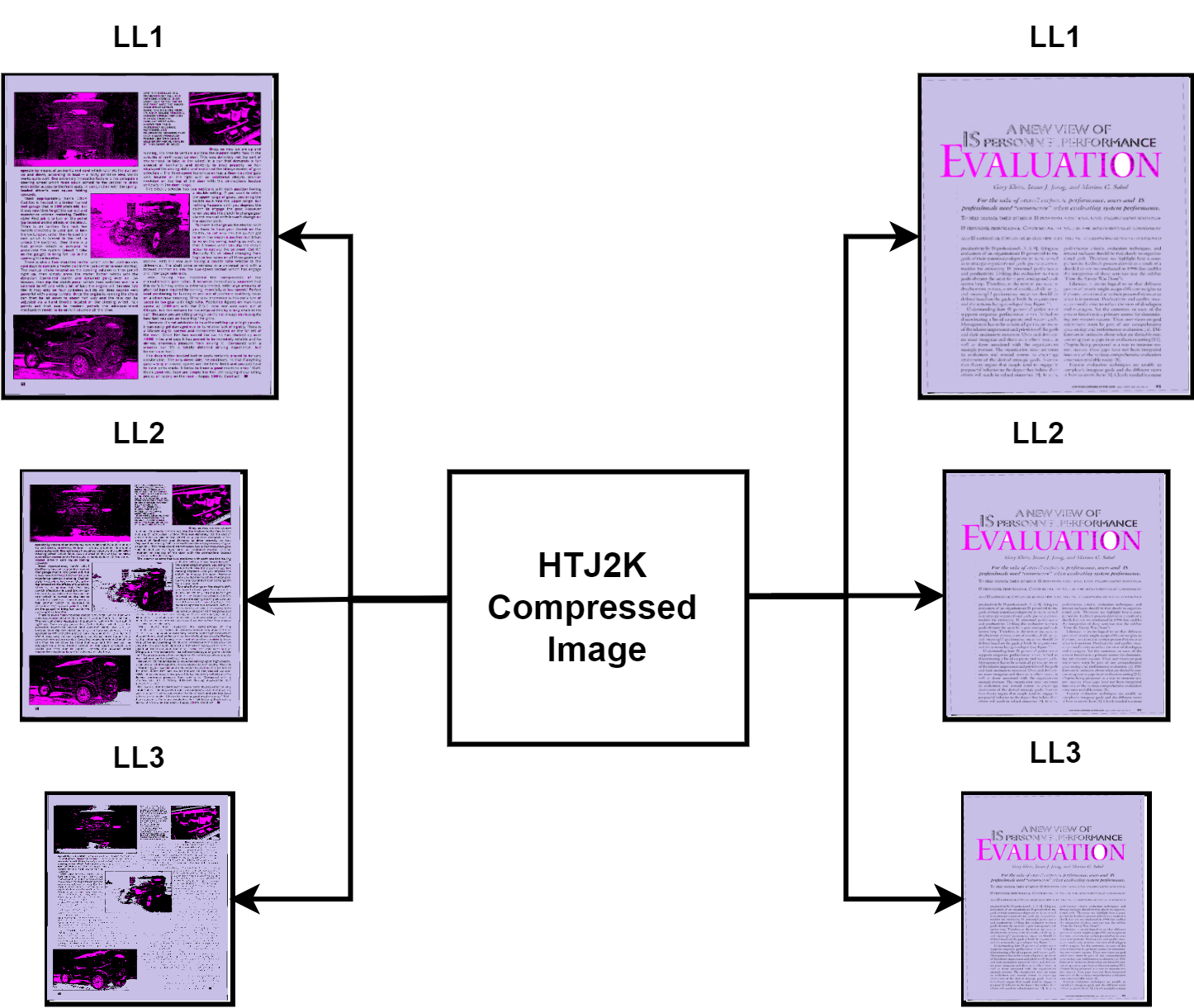}} %\vspace*{8pt}
		\caption{ Three level decomposition of documents containing text and non-text, and only text, illustrated using DWT decomposition, where three level LL subbands are used by the proposed DWT-CompCNN model.}
		\label{ll}
	%\end{center}
\end{figure*}

\section{Proposed Approach}\label{sec2}
This section explains the proposed approach for classification of document images at different resolutions of DWT directly from compressed HTJ2K images using a deep learning model. The objective here is to set label to compressed images of different classes by partially decompressing at different resolutions of DWT, and make the whole process of classification very efficient in terms of accuracy, time and space. The proposed classification model termed as DWT-CompCNN is illustrated with a block diagram as shown in  Fig.\ref{proposed model}.

The DWT-CompCNN takes the HTJ2K compressed images as input, and partially decompress them to extract the DWT coefficients up to $L=3$ resolutions for feature extraction using the different convolutional layers. The model is trained with all three resolutions of DWT where $1^{st}$ resolution contains coarsest level features, $2^{nd}$ resolution has fine level features, and the $3^{rd}$ resolution has the finest detail of the image. The process of DWT decomposition at all three resolutions is illustrated in the Fig.\ref{ll} with the help documents containing text and non-text, and only text. From the figure it is very clear that smaller objects like text require higher resolution, and non-text objects can be processed at lower resolutions. During the testing phase, the compressed images in the validation set are given to the proposed model, where class label is assigned using fully connected layers with the help of features learning that happened during the training phase of DWT-CompCNN. 

\begin{figure*}[h]
	\begin{center}
		\centering{\includegraphics[width=12cm,height=7cm]{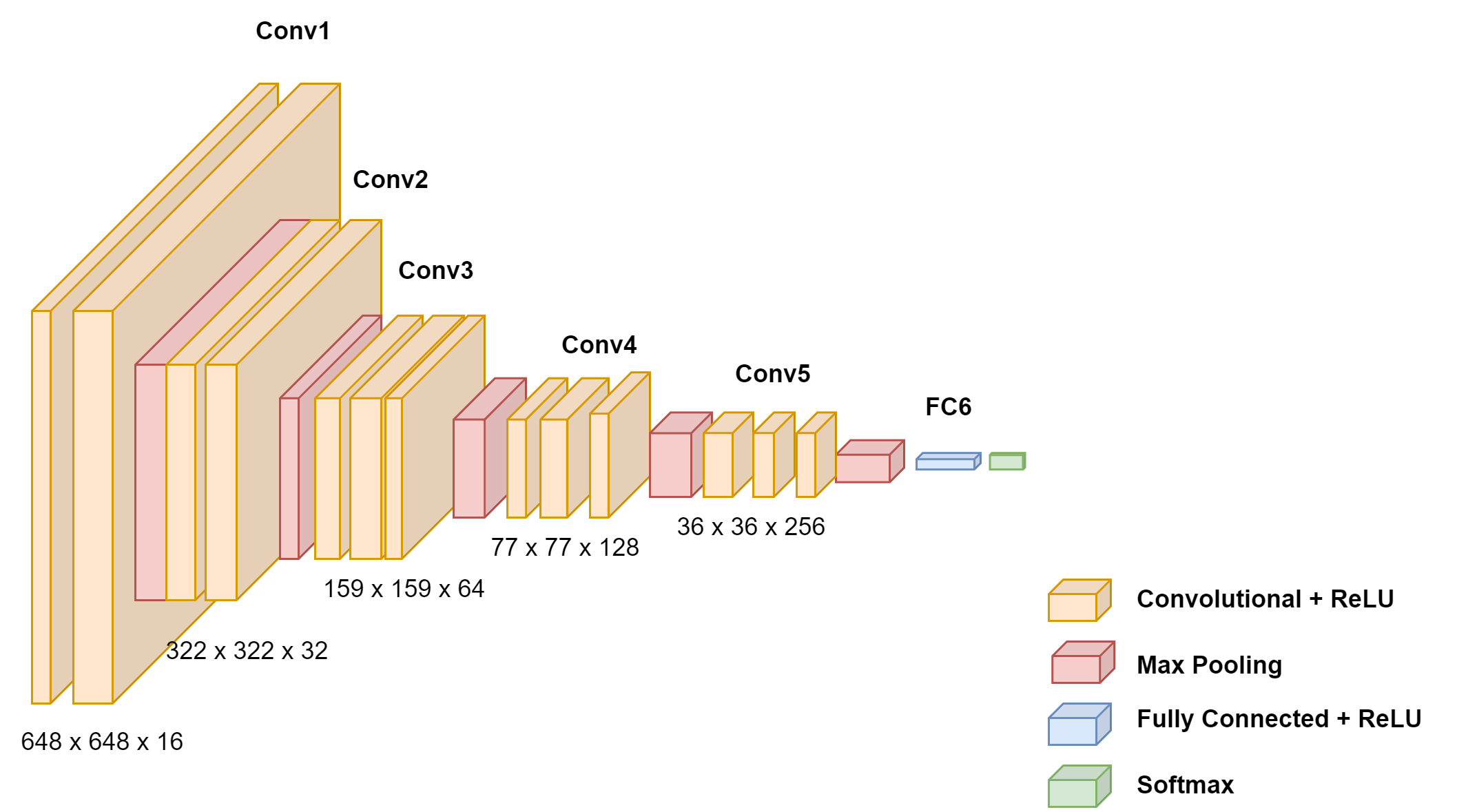}} \vspace*{8pt}
		\caption{ The detailed architecture of proposed DWT-CompCNN for HTJ2K compressed document images.}
		\label{CNN model}
	\end{center}
	
\end{figure*}

The architecture of the proposed DWT-CompCNN model shown in Fig.\ref{CNN model} is inspired from the existing VGG16 \cite{zhang2015accelerating}, which is a pretrained model on ImageNet dataset having a large number of classes. VGG16 has 16 layers with a filter size starting from 64 up to 512. Due to large filter size at the beginning layers, VGG16 could not extract finer details, and hence the performance on the compressed images were poor. This is because the compressed images have compact information and less redundancy, hence a small size filter is required to learn features from the compressed images. Also, VGG16 was designed for a large number of classes, however in our problem, as we have less number of classes (10 and 16 classes for Tobacco-3482 and RVL-CDIP respectively), the number of layers have to be reduced to improve the computational efficiency of the proposed model. Therefore, the proposed DWT-CompCNN comprises five convolutional layers with filter sizes of 16, 32, 64, 128, and 256 consecutively for each increasing layer to improve learning from compressed images. Also in JPEG2000, codeblock is the smallest geometric structure that can be equal to the precinct size which is 16x16. Therefore filter size 16 in the first layer of DWT-CompCNN facilitates good feature extraction.
Each convolutional layer has a different receptive scale factor \cite{lecun1998gradient}, which shows, how each node of feature maps are connected to different regions of the image. A smaller receptive scale factor contributes to the finer features, while a larger receptive scale factor contributes to the coarser features. 
The feature maps in the DWT-CompCNN contain exhaustive and structured information, and they are integrated to perform the task of classification. Conv1 layer uses the 16x16 filter and Conv2 layer uses 32x32 filter for extracting lower-level and regional features. The Conv5 layer with a filter size of 256x256 extracts the higher level and global features of images. The feature maps of Conv3 layer are more sensitive to the edges, but Conv4 layer and Conv5 layer feature maps emphasize textual regions. The features such as shapes, edges and texture are extracted in these five convolutional layers. After flattening the feature maps of the final convolutional layer, they are pooled using max pooling and then passed to the fully connected layer FC6, which is similar to that of VGG16 architecture. Fully connected layers enable the combining of information from each input and each output class, which helps the model make a decision to assign a proper class to the image based on the extracted information.
The max pooling layer is used to reduce the dimension and computation. The rectified linear unit (ReLU) activation function is used in intermediate layers to speed up the training of the proposed model and introduce non-linearity. The softmax activation function is introduced at end of the fully connected layer, to give the probability score to the images. Dropouts remarkably decreases the overfitting and also increase the learning speed of the model, by omitting the training nodes on training data. Dropouts of 10, 15, 20, 25, and 30 percent are implemented after each convolutional layer to make the proposed model efficient. The loss function is used to assess the prediction error and inconsistency between the actual and predicted class of the given image. The model's efficiency increases as the loss function's value decreases. The classification loss function for the proposed model is categorical cross entropy, which is used as training and validation loss. The loss function is used for adjusting the weight of the model while training to reduce the loss and make the proposed model efficient. For the $N$ number of image classes, $a_i$ is the actual class for the given image in the dataset and $p_i$ is the softmax probability predicted for the $i^{th}$ class. The categorical cross entropy (CE) loss function is defined in the Equation (\ref{eq5})\\
\begin{equation}
CE_{(a_{i}, p_{i})} = - \frac{1}{N}\left[ \sum_{i=1}^{} \left[ a_{i}\log\left( p_{i} \right) + \left( 1 - a_{i} \right) \log\left( 1 - p_{i} \right) \right] \right]
\label{eq5}
\end{equation}
\\
The experiments are performed on the proposed approach for the classification of document images at different resolutions for different learning rates, batches, and epochs to get better results, and are discussed in the next section.

\section{Experiments and Results}

\begin{figure*}
%	\begin{center}
		\centering{\includegraphics[width=\textwidth,height=8cm]{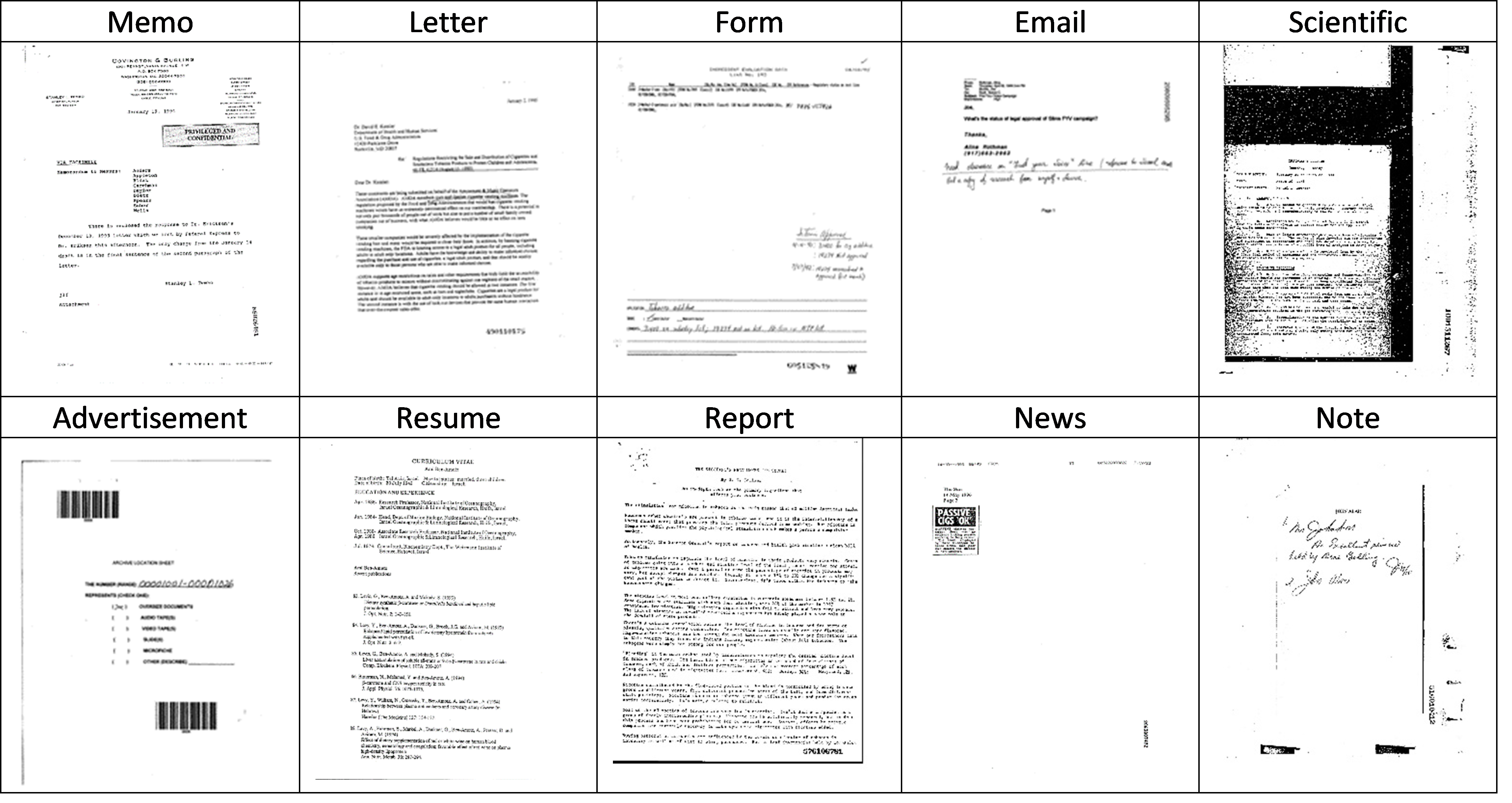}} \vspace*{2pt}
		\caption{ Sample document images of Tobacco-3482 dataset of 10 different classes \cite{lewis2006building}}
		\label{Tobacco_3482_sample}
%	\end{center}
\end{figure*}

\begin{figure*}
	%\begin{center}
		\centering{\includegraphics[width=\textwidth,height=5cm]{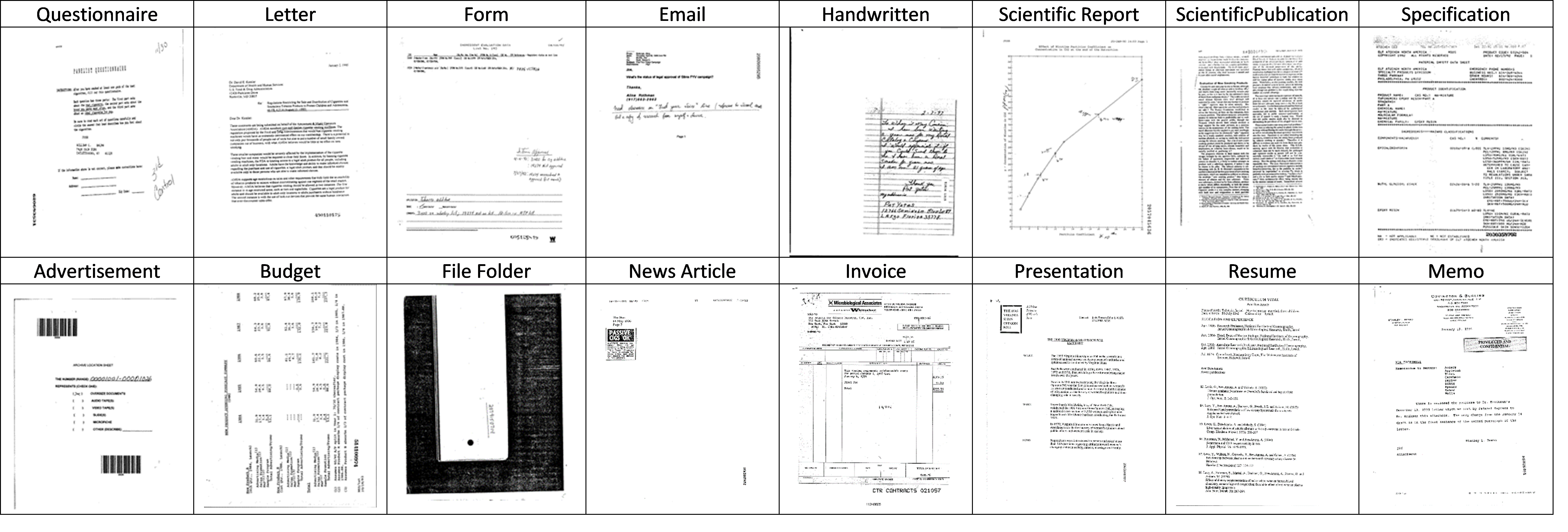}} \vspace*{2pt}
		\caption{ Sample document images of RVL-CDIP dataset of 16 different classes \cite{harley2015icdar}}
		\label{RVL_sample}
	%\end{center}
	
\end{figure*}

\subsection{Dataset and Experimental Protocol}
The proposed DWT-CompCNN model's performance is evaluated on the two standard document image datasets, Tobacco-3482 and RVL-CDIP respectively using the HTJ2K compressed version of images.
The first dataset is Tobacco-3482 \cite{lewis2006building} which consists of 3482 images, with 10 different classes (memo, resume, news, scientific, letters, email, report, advertisement, note, form). The sample images of the Tobacco-3482 dataset for each class are shown in  Fig.\ref{Tobacco_3482_sample}.
The second dataset is RVL-CDIP dataset \cite{harley2015icdar} which consists of 400,000 grayscale images, with 16 different classes (letter, form, email, handwritten, advertisement, scientific report, scientific publication, specification, file folder, news article, budget, invoice, presentation, questionnaire, resume, memo). Each class has 25,000 images. The sample images of the RVL-CDIP dataset for each class are shown in Fig.\ref{RVL_sample}. 
The HTJ2K \cite{fang2021development} compressed version of the above two datasets are used for experimental studies.

Both datasets are divided into 80\% as training sets and 20\% of training datasets  as validation sets. The experimental results are shown in Table-1 (Tobacco-3482) and Table-5 (RVL-CDIP). The experiments on the model are carried out at a learning rate of 0.001 using the adaptive moment estimation \cite{kingma2014adam}. It helps the model to learn significant features at different epochs by discriminating the insignificant features. Furthermore, experiments were conducted with learning rates ranging from 0.1 to 0.0001 in order to attain accurate performance.
The proposed DWT-CompCNN model is evaluated on the following parameters: Classification Accuracy, Computational time, and Computational space. All the experiments are conducted on Nvidia Tesla-V100 having a 40,960 CUDA core. The MATLAB implementation of HTJ2K version \cite{watanabe2019matlab} is used in this research work.
%Nvidia Tesla-V100 having a 40,960‬ CUDA core. The MATLAB implementation of HTJ2K version \cite{watanabe2019matlab} is used in this research work.

The Classification Accuracy $(CA_{MC})$ for multi class is defined in the Equation (\ref{eqCA}), where $N$ is number of image classes, $TP_i$, where the given image is correctly predicted class for the positive label, $TN_i$, where the given image is correctly predicted class for the negative label, $FP_i$, where the given image is predicted to a positive label but actual label is negative and $FN_i$ where the given image is predicted to a negative label but actual label is positive. The training accuracy is the number of correctly predicted images to the number of samples used during the training phase. The validation accuracy for the model is used to evaluate the performance of the model during the validation phase.\\
\begin{equation}
CA_{MC}= \dfrac{\sum _{i=1} ^N \dfrac{TP_i + TN_i}{TP_i + TN_i + FP_i + FN_i}}{N}
\label{eqCA}
\end{equation}

\subsection{Experiments on Tobacco-3482 Dataset}
The proposed approach's classification accuracies, computational time and computational space are discussed in this subsection with Tobacco-3482 dataset.
The Table \ref{tab2} shows the performance of the proposed DWT-CompCNN model on the Tobacco-3482 dataset, in terms of classification accuracy and validation loss at three different resolutions in the HTJ2K compressed domain. The classification accuracy of the model for the finest level (3$^{rd}$ resolution) is 92.04\%, for fine level (2$^{nd}$ resolution) is 89.06\%, and for the coarsest level (1$^{st}$ resolution) is 78.36\%. Since there are no state-of-the art work related to classification of document images in HTJ2K compressed domain, the performance of the proposed model is compared with existing literature in the uncompressed domain. Table \ref{pc_tobacco} shows that the VGG16 and ResNet50 have obtained an accuracy of 90.18\% and 88.89\% respectively in the pixel domain, which is much lower than that of our proposed model. The table shows that the proposed DWT-CompCNN model has achieved better accuracy for classifying the images directly from the compressed domain in comparison to existing works of the pixel domain.\\ 
To asses the time computing performance of the proposed DWT-CompCNN model, we have used the computational time of 25 document images decompressed at all three resolutions and full decompression as shown in the Table \ref{timetab_tobacco}. The computation time ($CT_s$) includes the decompression time ($DT_s$) of the LL subband of the dataset and the  classification time ($CLT_s$), to classify the images in the predefined different classes, which is formulated in the Equation (\ref{eqCT}): 
\begin{equation}
   CT_{s} = DT_{s} + CLT_{s} 
   \label{eqCT}
\end{equation}

Table \ref{timetab_tobacco} shows that computational time decreases with decrease in resolution, also the computational time for the fully decompressed image is much higher in comparison with the decompressed image at  3$^{rd}$ resolution. The highest level resolution (third resolution) takes 1235.62 sec computation time for classifying images, where as the second resolution takes 727.31 sec and the first resolution takes 489.18 sec. The finest level (3$^{rd}$ resolution) is taking more computational time as compared to the other two levels but it is less than a full decompressed image, which is 2353.24 sec. The Speedup is the another performance metric, which is used to evaluate the proposed model.
Speedup ($S_{L}$) is defined in the Equation (\ref{eqspeedup}) as the ratio of computational time of fully decompressed image ($CT_{I_{fd}}$) to computational time of partially decompressed image ($CT_{I_{pd_{L}}}$) at the resolution level ($L$).
\begin{equation}
 S_{L} = \dfrac{CT_{I_{fd}}}{CT_{I_{pd_{L}}}}  
 \label{eqspeedup}
\end{equation}
In Table \ref{timetab_tobacco}, speedup for the highest resolution (3$^{rd}$) is 1.90, which increases to 3.24 for the the second resolution and 4.81 for the first resolution. Here the model's performance is much better for partially decompressed images than the fully decompressed image.
\subsection{Experiments on RVL-CDIP Dataset}
 Table \ref{tab1} shows the performance of the proposed DWT-CompCNN model on the RVL-CDIP dataset. The classification accuracy of the model for the finest level ($3^{rd}$ resolution) is 98.94\%, for finer level ($2^{nd}$ resolution) is 97.79\%, and for the coarsest level ($1^{st}$ resolution) is 88.61\%. The Table  \ref{pc_RVL} shows the comparison of the proposed model with the previous works in the pixel domain. Our model outperforms the existing classification techniques in the uncompressed domain as shown in Table  \ref{pc_RVL}.\\
 
Table \ref{timetab_rvl} shows the computational time for the classification of images at three different resolutions. The computational time reported for 25 documents at third resolution is 25316.72 sec, at second resolution is 25285.35 sec and for the first resolution is 25033.71 sec. The computational time for fully decompressed image is 27961.42 sec which is greater than the partially decompressed image at third resolution. There will be a significant amount of time saving when the whole dataset is taken into account. The table also indicates the speedup of the proposed DWT-CompCNN model, which is 1.10 at 3$^{rd}$ resolution, 1.11 at $2^{nd}$ and 1.11 at 1$^{st}$ resolution, which shows the model's speed is similar for all three resolutions. The speedup for Tobacco-3482 dataset is much more than the RVL-CDIP dataset, because Tobacco-3482 dataset contains high resolution images, and RVL-CDIP dataset images are already in the low resolution. Hence it can be concluded that DWT-CompCNN model is more suitable for high resolution images taking the advantage of different resolutions of DWT decomposition.\\

\begin{table*}[!t]
\centering
\caption{Classification accuracy for the proposed DWT-CompCNN model for Tobacco-3482 dataset for three different resolutions from highest to lowest in the compressed domain of HTJ2K with 100 epochs}\label{tab2}
\begin{adjustbox}{width=\textwidth}{
\begin{tabular}{ccccc}
\hline
\hline
\textbf{Resolutions} & \textbf{Training Accuracy} & \textbf{Training Loss} & \textbf{Validation Accuracy} & \textbf{Validation Loss}  \\\hline \hline
3 & 95.68\% & 0.1539 & 92.04\% & 0.4610  \\
2 & 93.18\% & 0.2020 & 89.66\% & 0.8610  \\
1 & 91.82\% & 0.2453 & 78.36\% & 2.0091   \\
\hline
\hline
\end{tabular}
}
\end{adjustbox}
\end{table*}

\begin{table*}
\begin{center}
\centering
   \caption{Performance comparison of existing document image classification algorithms in the uncompressed domain with the proposed DWT-CompCNN model for Tobacco-3482 dataset }\label{pc_tobacco}
 \begin{adjustbox}{width= \textwidth}{
    \begin{tabular}{l|p{3cm}|c}
     \hline \hline
      \textbf{Methods} & \textbf{Dataset} & \textbf{Accuracy}\\ \hline \hline
        Kumar et al \cite{kumar2014structural} & \multirow{10}{3cm}{Uncompressed Tobacco-3482 Dataset} & 43.80 \% \\
        Kang et al \cite{kang2014convolutional} & & 65.37 \% \\
        Afzal et al \cite{afzal2015deepdocclassifier} & & 77.60 \% \\
        Noce et al \cite{afzal2015deepdocclassifier} &  & 79.80 \% \\
        Harley et al \cite{harley2015evaluation} & & 79.90 \% \\
       Sarkhel and Nandi et al [\cite{sarkhel2019deterministic}] & & 82.78 \% \\
        Kolsch et al \cite{kolsch2017real} & & 83.24 \% \\
        Afzal et al \cite{afzal2017cutting} & & 91.13 \% \\
        Kanchi et al \cite{kanchi2022emmdocclassifier} & & 90.30 \% \\
        VGG16 & & 90.18\% \\
        ResNet50 & & 91.13\% \\
       \hline
        \textbf{Proposed DWT-CompCNN} & HTJ2K compressed Tobacco-3482 Dataset & 92.04\% \\
      \hline \hline
      \end{tabular}
      }
\end{adjustbox}
\end{center}
\end{table*}

\begin{table*}[!t]
\centering
\caption{Computational time comparison of proposed DWT-CompCNN model with uncompressed version and three different resolutions for Tobacco-3482 dataset (in seconds)}\label{timetab_tobacco}
\begin{adjustbox}{width=\textwidth}{
\begin{tabular}{c|c|c|c|c|c|c}
\hline \hline
\multicolumn{6}{c} {\textbf{Tobacco-3482}} \\ \hline
Decompression type & Resolution & No. of Images &  Decompression Time & Classification Time & Computational Time & Speedup\\ \hline
Full Decompression & Original Image & 25 & 1928.24 & 425 & 2353.24 & \\ \hline
\multirow{3}{3cm}{Partial Decompression} & Resolution 3 & 25 &  910.62 & 325 & 1235.62 & 1.90 \\
& Resolution 2 & 25 & 452.31 & 275 & 727.31 & 3.24 \\
& Resolution 1 & 25 & 239.18 & 250 & 489.18 & 4.81 \\
\hline \hline
\end{tabular}
}
\end{adjustbox}
\end{table*}

\begin{table*}
\centering
\caption{Space complexity at different resolutions used by proposed DWT-CompCNN model}\label{spacetab}
\begin{adjustbox}{width=\textwidth}{
\begin{tabular}{ccc}
\hline \hline
\multicolumn{1}{c}{} &
\multicolumn{1}{c}{\textbf{ Tobacco-3482}}&
\multicolumn{1}{c}{\textbf{RVL-CDIP}} \\ \hline
Resolutions & Computational space & Computational space\\ \hline \hline
Uncompressed Image & 1200x1575 & 762x1000  \\
Resolution 3 & 600x788 & 377x500 \\
Resolution 2 & 300x394 & 189x250 \\
Resolution 1 & 216x287 & 94x125 \\
\hline \hline
\end{tabular}
}
\end{adjustbox}
\end{table*}

\begin{figure}[h]
	\centering{\includegraphics[width=7.5cm,height=5cm]{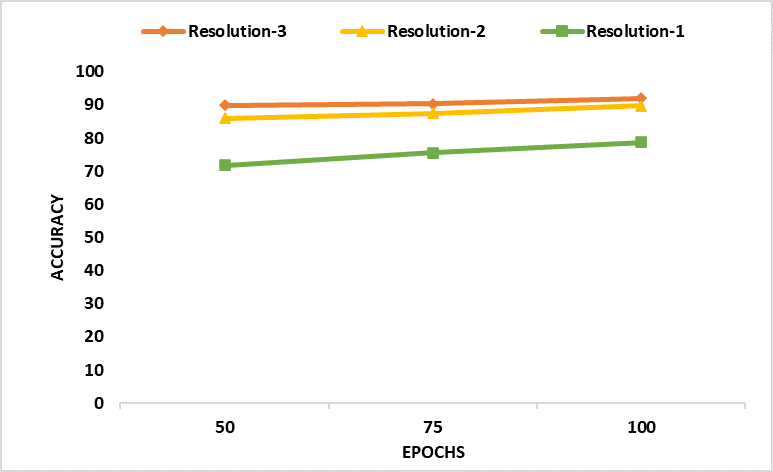}}
		\caption{ Validation Accuracy at different resolutions for batch size of 32 at different epochs for Tobacco-3482 dataset} 
		\label{fig6}
\end{figure}
\begin{table*}
\begin{center}
%\begin{subtable}
\centering
\caption{Classification accuracy for the proposed DWT-CompCNN model for RVL-CDIP dataset for three different resolutions from highest to lowest in compressed domain of HTJ2K with 100 epochs}\label{tab1}
\begin{adjustbox}{width=\textwidth}{
\begin{tabular}{ccccc}
\hline \hline
\textbf{Resolutions} & \textbf{Training Accuracy} & \textbf{Training Loss} & \textbf{Validation Accuracy} & \textbf{Validation Loss}  \\\hline \hline
3 & 99.96\% & 0.0212 & 98.94\% & 0.2016  \\
2 & 98.69\% & 0.0461 & 97.79\% & 0.4511  \\
1 & 98.28\% & 0.0626 & 88.61\% & 0.5921  \\
\hline
\hline
\end{tabular}
}
\end{adjustbox}
\end{center}
\end{table*}
%\end{subtable}
%\begin{subtable}
\begin{table*}
\begin{center}
\centering
 \caption{Performance comparison of existing document image classification algorithms in the uncompressed domain with the proposed DWT-CompCNN model for RVL-CDIP dataset }\label{pc_RVL}
 \begin{adjustbox}{width=\textwidth}{
    \begin{tabular}{l|p{3cm}|c}
     \hline \hline
      \textbf{Methods} & \textbf{Dataset} & \textbf{Accuracy}\\ \hline \hline
      
        Harley et al \cite{harley2015evaluation}& \multirow{6}{3cm}{Uncompressed RVL-CDIP Dataset} & 89.90 \% \\
        Csurka et al \cite{csurka2016right} & &  90.70 \% \\
        Afzal et al \cite{afzal2017cutting} & & 90.97 \% \\
        Tensmeyer and Martinez et al \cite{tensmeyer2017analysis} & & 91.03 \% \\
        Das et al \cite{das2018document} & & 92.21 \% \\
        Ferrando et al \cite{ferrando2020improving} & & 92.31 \% \\
        Bakkali et al \cite{bakkali2020visual} & & 97.05 \% \\
       \hline
        \textbf{Proposed DWT-CompCNN} & HTJ2K compressed RVL-CDIP Dataset & 98.94 \% \\
      \hline \hline
      \end{tabular}  
      }
\end{adjustbox}
%\end{subtable}
\end{center}
\end{table*}

\begin{table*}[!t]
\centering
\caption{Computational time comparison of proposed DWT-CompCNN model with uncompressed version and three different resolutions for RVL-CDIP dataset (in seconds)}\label{timetab_rvl}
\begin{adjustbox}{width=\textwidth}{
\begin{tabular}{c|c|c|c|c|c|c}
\hline \hline
\multicolumn{6}{c}{\textbf{RVL-CDIP}} \\ \hline
Decompression type & Resolution & No. of Images &  Decompression Time & Classification Time & Computational Time & Speedup\\ \hline
Full Decompression & Original Image & 25 & 36.42 & 27925 & 27961.42 & \\ \hline
\multirow{3}{3cm}{Partial Decompression} & Resolution 3 & 25 &  16.72 & 25300 & 25316.72 & 1.10 \\
& Resolution 2 & 25 & 10.35 & 25275 & 25285.35 & 1.10 \\
& Resolution 1 & 25 & 8.71 & 25025 & 25033.71 & 1.11\\
\hline \hline
\end{tabular}
      }
\end{adjustbox}
\end{table*}

\subsection{Further Discussions}
\begin{figure}[h]
	\centering{\includegraphics[width=7.5cm,height=5cm]{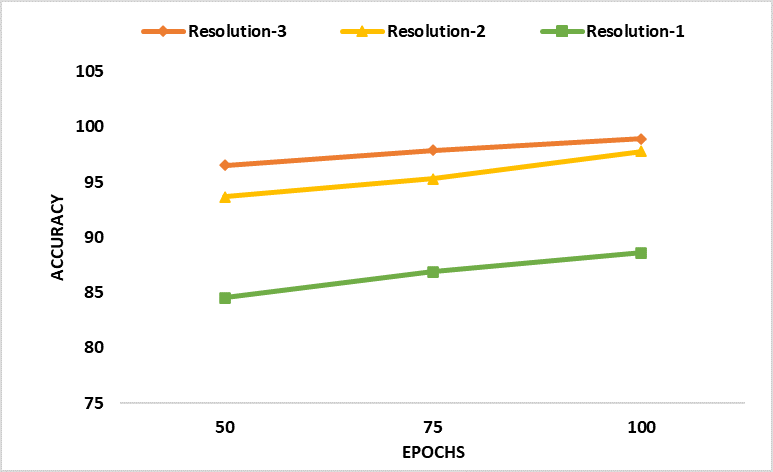}}
		\caption{ Validation Accuracy at different resolutions for batch size of 32 at different epochs for RVL-CDIP dataset} 
		\label{fig5}
\end{figure}

By analyzing the results of the proposed DWT-CompCNN model on the two different datasets, Tobacco-3482 and RVL-CDIP, we can interpret that the proposed DWT-CompCNN model significantly reduces the computational time.
The graphs in Fig.\ref{fig6} and Fig.\ref{fig5} show the validation accuracy of the proposed model at three different resolutions for the two datasets. It represents that the classification accuracy of the proposed DWT-CompCNN model increases for the higher resolutions as there is less details stored in the LL subband of lower resolutions. 

\begin{figure*}
	\begin{center}
		\centering{\includegraphics[width= \textwidth,height=14cm]{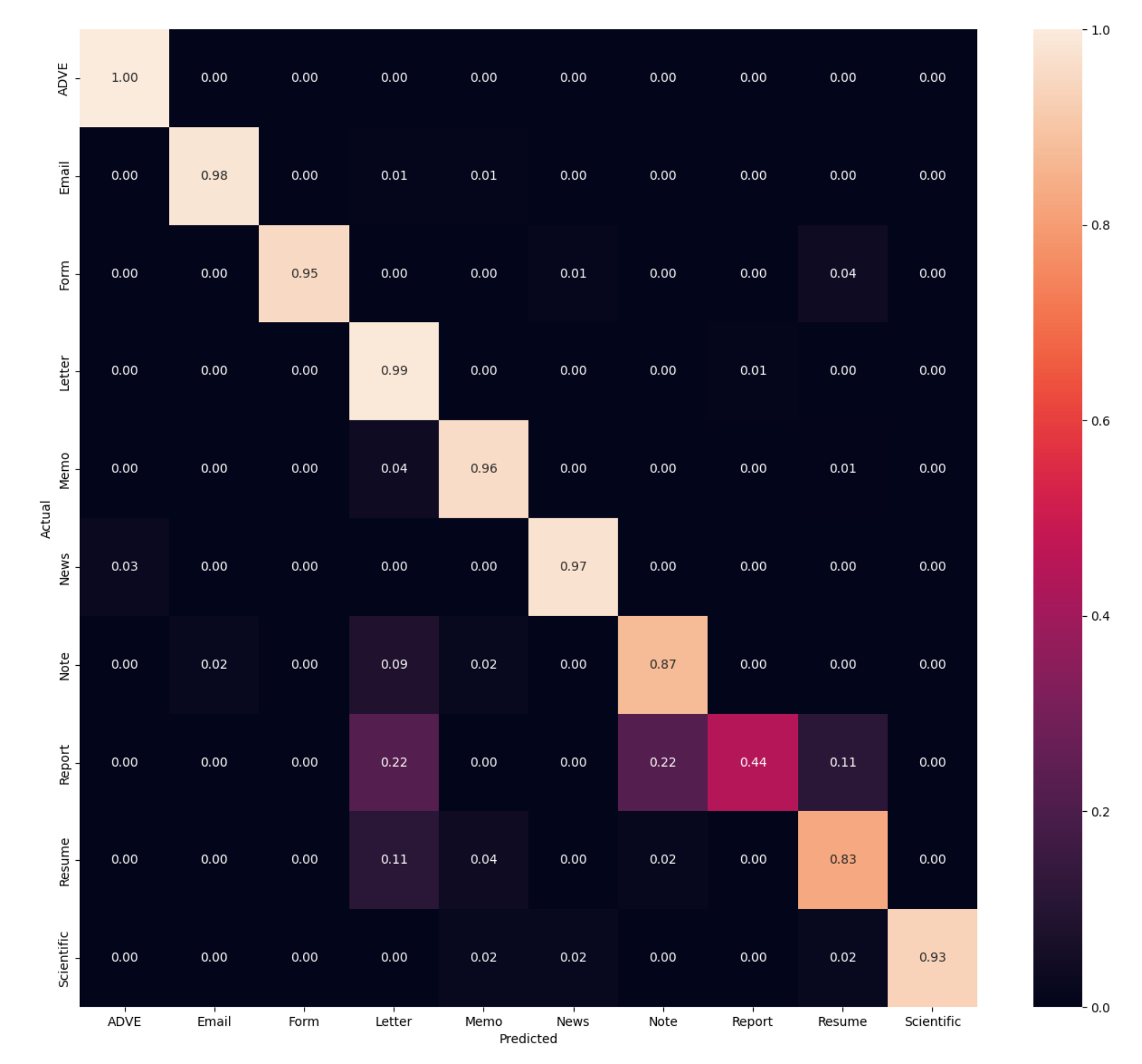}}
		\caption{ Performance of proposed DWT-CompCNN model is illustrated through confusion matrix. The diagonal number represents the percentage wise accuracy for that particular class, and off-diagonal shows the missclassified, for the resolution-3 (highest) of Tobacco-3482 dataset.}
		\label{conf_Tobacco_3482}
	\end{center}
\end{figure*}

\begin{figure*}
	\begin{center}
		\centering{\includegraphics[width=\textwidth,height=15cm]{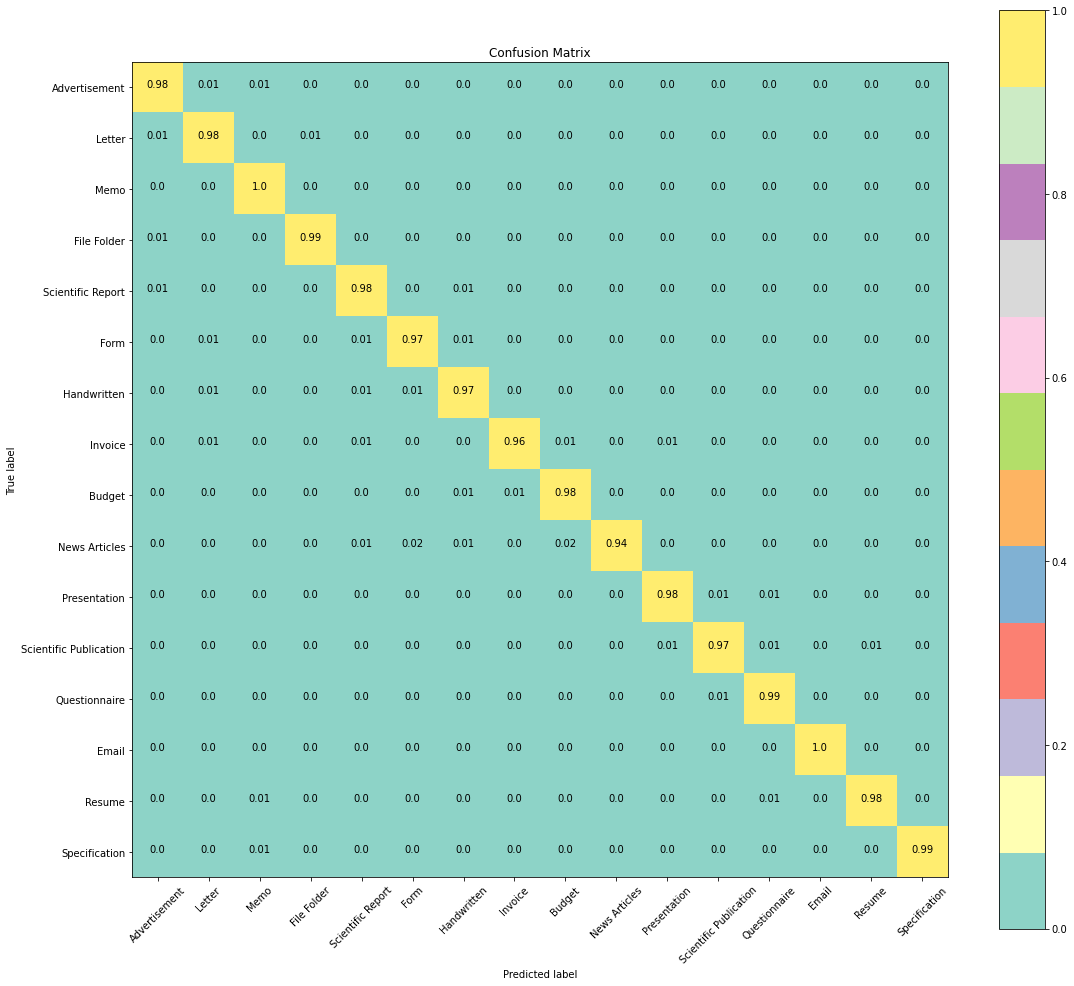}}
		\caption{ Performance of proposed DWT-CompCNN model is illustrated through confusion matrix. The diagonal number represents the percentage wise accuracy for that particular class, and off-diagonal shows the missclassified, for the resolution-3 (highest) of RVL-CDIP dataset.}
		\label{conf_rvl}
	\end{center}
\end{figure*}

In Fig. \ref{conf_Tobacco_3482}, we can visualize the accuracy of different classes for the Tobacco-3482 dataset. It is observed that our proposed model has performed better in most of the classes.  It is able to achieve accuracy above 90\% in most of the classes and hence the overall accuracy reaches to 92.04\%. In the case of the “News” class which has the lowest number of samples in the dataset, an accuracy of 97\% is still observed. The class "Report" is the most misleading one, as it has a classification accuracy of 44\%, which is misclassified with the classes “Letter”, “Note” and “Resume”. This is because they are overlapping classes having similar structure and hence the misclassification is observed. 
Similarly, in Fig. \ref{conf_rvl}, the confusion matrix of the RVL-CDIP dataset is shown. The proposed model has achieved an accuracy of 99\% in most of the classes. The most misleading class here is “News Articles”, which resulted in accuracy of 94\%, and here the misclassifications were observed with the classes- Scientific Report (0.01\%), Handwritten (0.01\%), Budget (0.02\%), and Form (0.02\%). The overall accuracy with DWT-CompCNN  on the RVL-CDIP dataset is 98.94\%. Also, Table \ref{tabconftob} and Table \ref{tabconfrvl} show the performance of the proposed model for other metrics such as precision, recall, and F1 score \cite{kumar2014structural} for different classes.

The proposed DWT-CompCNN was experimented with different training sizes, and the performance observed is reported in Table \ref{trainingtab}. It was concluded that variations in training size have a minor impact on both the training and validation accuracy. The table shows that with 40\% of the Tobacco-3482 training dataset, the proposed model is not able to cross 90\% of validation accuracy, but as the training data increases, there is an increase in validation accuracy. Similarly, the RVL-CDIP dataset, achieves good accuracy as the training dataset is increased. This is because the RVL-CDIP dataset has a larger number of samples for each class compared to Tobacco-3482.

The space complexity for the proposed model is described in the Table \ref{spacetab}, which shows that computational space is reduced significantly at each resolution level for both the datasets. This is because of dyadic decomposition used by DWT, where the LL subband is decomposed in recursive manner. At the first level decomposition, $M$ buffer lines are  required for vertical decomposition. Further at the second level M buffer lines are again required, but for $Z/2$ coefficients. Similar memory is required for further decompositions. Here, the dimension of the image is reduced by 50\% at each level, the memory requirement is also decreased by 50\% each time. The buffer required for each level of coefficients is mentioned in the Equation (\ref{dyd}).\\
\begin{equation}
    3. (2^{L-l} -1)S.2^{-l-1}Z
    \label{dyd}
\end{equation}
where $L$ is number of decomposition levels (3 levels in the proposed approach), $l$ is first decomposition level, $S$ is the required time for first level decomposition and $Z$ is the amount of space occupied ($Z$ is width of the image).
And the total memory ($TM$) required for coefficients and synchronization delay increases as we move to higher resolution, so Equation (\ref{mem}) represents the total memory requirement for the $L$ level decomposition used by DWT:\\
\begin{equation}
    TM^{2Dy}_{L} = 3 \sum_{L-1}^{l=0} (2^{L-l}-1)SZ2^{-l-1} = (2.2^{L} + 2^{-L} -3)ZS
    \label{mem}
\end{equation}
Equation (\ref{mem}) shows the memory requirement at each higher decomposition level increases, where the synchronization buffer increases exponentially but the filtering buffer bound is $2MZ$. The proposed approach works on an HTJ2K compressed version of the image, so it requires less memory as compared to the requirement of the original version of images.\\

Overall, the proposed DWT-CompCNN model achieved better accuracy in classifying the images directly in the HTJ2K compressed domain.  Experimentally the model was proved to have high accuracy with reduced  computational time and space.
%\lipsum[3-5]
\begin{table*}
\centering
\caption{Training and Validation accuracies by variation in Training Set Sizes}\label{trainingtab}
\begin{adjustbox}{width=\textwidth}{
\begin{tabular}{c|c|c|c|c}
\hline \hline
\multicolumn{1}{c}{} &
\multicolumn{2}{c}{\textbf{ Tobacco-3482}}&
\multicolumn{2}{c}{\textbf{RVL-CDIP}} \\ \hline
Training Size & Training Accuracy  & Validation Accuraccy & Training Accuracy  & Validation Accuracy\\ \hline \hline
40\% & 96.89\% & 89.43\% & 99.98\% & 96.42\%  \\
50\% & 96.05\% & 89.71\% & 99.97\% & 96.67\%  \\
70\% & 95.87\% & 91.89\% & 99.96\% & 98.80\%  \\
80\% & 95.68\% & 92.04\% & 99.96\% & 98.94\%  \\
\hline \hline
\end{tabular}
}
\end{adjustbox}
\end{table*}
\begin{table*}
\caption{Performance of the proposed DWT-CompCNN model for 16 different classes of Tobacco-3482 dataset at 3$^{rd}$ resolution}\label{tabconftob}
\begin{adjustbox}{width=\textwidth,height=3cm}{
\begin{tabular}{lcccc}
\hline
\hline
\textbf{Classes} & \textbf{Precision} & \textbf{Recall} & \textbf{F1 Score} & \textbf{Accuracy}\\ \hline \hline
Advertisement (ADVE) & 1.0 & 0.25 & 0.40 & 99.67\%  \\
Email & 0.98 & 0.98 & 0.98 & 99.56\%  \\
Form & 0.95 & 1.0 & 0.97 & 99.44\%   \\
Letter & 0.99 & 0.68 & 0.80 & 94.66\% \\
Memo & 0.96 & 0.91 & 0.93 & 98.44\% \\
News & 0.97 & 0.98 & 0.97 & 99.44\% \\
Report & 0.87 & 0.78 & 0.82 & 95.88\% \\
Resume & 0.44 & 0.98 & 0.61 & 93.77\% \\
Scientific & 0.83 & 0.83 & 0.83 & 96.22\% \\
Note & 0.94 & 1.0 & 0.97 & 99.33\% \\
\hline
\hline
\end{tabular}
}
\end{adjustbox}
\end{table*}

\begin{table*}
\caption{Performance of the proposed DWT-CompCNN model for 16 different classes of RVL-CDIP dataset at 3$^{rd}$ resolution}\label{tabconfrvl}
\begin{adjustbox}{width=\textwidth,height=5cm}{
\begin{tabular}{lcccc}
\hline
\hline
\textbf{Classes} & \textbf{Precision} & \textbf{Recall} & \textbf{F1 Score} & \textbf{Accuracy}\\ \hline \hline
Advertisement & 0.98 & 0.97 & 0.98 & 99.64\%  \\
Letter & 0.98 & 0.96 & 0.97 & 99.57\%  \\
Memo & 1.0 & 0.25 & 0.40 & 99.79\%   \\
File Folder & 0.99 & 0.99 & 0.99 & 99.86\% \\
Scientific Report & 0.98 & 0.96 & 0.97 & 99.57\% \\
Form & 0.97 & 0.97 & 0.97 & 99.57\% \\
Handwritten & 0.97 & 0.96 & 0.97 & 99.50\% \\
Invoice & 0.96 & 0.99 & 0.97 & 99.64\% \\
Budget & 0.98 & 0.97 & 0.98 & 99.64\% \\
News Articles & 0.94 & 1.0 & 0.97 & 99.57\% \\
Presentation & 0.98 & 0.98 & 0.98 & 99.71\% \\
Scientific Publication & 0.97 & 0.98 & 0.97 & 99.64\% \\
Questionnaire & 0.99 & 0.97 & 0.98 & 99.71\% \\
Email & 1.0 & 1.0 & 1.0 & 100\% \\
Resume & 0.98 & 0.99 & 0.98 & 99.79\% \\
Specification & 0.99 & 1.0 & 0.99 & 99.93\% \\
\hline
\hline
\end{tabular}
}
\end{adjustbox}
\end{table*}
%\clearpage
\section{Conclusion}
In this paper, a deep learning model DWT-CompCNN was proposed to classify document images directly from the HTJ2K compressed document images. The proposed model was tested on two datasets, Tobacco-3482 and RVL-CDIP resulting in state-of-the-art accuracy with reduced computational time and space with the help of multiresolution feature of DWT in HTJ2K. The proposed model reported highest accuracy of 92.04\% and 98.94\% for Tobacco-3482 and RVL-CDIP datasets respectively. Also the highest Speedup of 4.81 was obtained for Tobacco-3482 dataset at resolution 1, and Speedup of 1.11 is obtained for RVL-CDIP dataset at resolution 1. The higher Speedup was observed in Tobacco-3482 dataset, because of high resolution images in the dataset. The proposed model can also find applications in remote sensing, medical images and any other type of high resolution images.

\clearpage
\bibliographystyle{IEEEtran}
\bibliography{references1}
%\bibliographystyle{sn-mathphys}
%\bibliography{references1.bib}% common bib file
%\input sn-article.bbl

%% Default %%%\input sn-sample-bib.tex

\end{document}